\documentclass[letterpaper]{article}
\usepackage[preprint]{aaai2027}
\usepackage[hyphens]{url}
\usepackage{graphicx}
\usepackage{natbib}
\usepackage{caption}
\usepackage{amsmath,amssymb}
\usepackage{algorithm}
\usepackage{algorithmic}
\usepackage{booktabs}
\usepackage{multirow}
\usepackage{xspace}

\newcommand{\method}{Memory-SAM\xspace}
\newcommand{\fg}{\mathrm{fg}}
\newcommand{\bg}{\mathrm{bg}}
\newcommand{\TopK}{\operatorname{TopK}}

\title{Memory-SAM: Human-Prompt-Free Tongue Segmentation via Retrieval-to-Prompt}
\author{
    Joongwon Chae\equalcontrib\textsuperscript{\rm 1,7},
    Lihui Luo\equalcontrib\textsuperscript{\rm 1},
    Yang Liu\textsuperscript{\rm 1},
    Xi Yuan\textsuperscript{\rm 2},
    Dongmei Yu\textsuperscript{\rm 4},
    Zhenglin Chen\textsuperscript{\rm 2},
    Runming Wang\textsuperscript{\rm 1},
    Ilmoon Chae\textsuperscript{\rm 7},
    Lian Zhang\textsuperscript{\rm 5},
    Peiwu Qin\corresponding\textsuperscript{\rm 6}
}
\affiliations{
    \textsuperscript{\rm 1}Tsinghua University, China\\
    \textsuperscript{\rm 2}The Fifth Affiliated Hospital of Wenzhou Medical University, China\\
    \textsuperscript{\rm 3}Shenzhen Traditional Chinese Medicine Hospital, China\\
    \textsuperscript{\rm 4}Wenzhou Medical University, China\\
    \textsuperscript{\rm 5}The First Hospital of Hebei Medical University, China\\
    \textsuperscript{\rm 6}Chinese Medicine Guangdong Laboratory/Hengqin Laboratory, China\\
    \textsuperscript{\rm 7}RatelSoft, Republic of Korea\\
    pwqin1979@gmail.com
}

\begin{document}
\maketitle

\begin{abstract}
Accurate tongue segmentation is required for quantitative tongue-image analysis, but automatic smartphone deployment remains difficult because SAM2 still depends on reliable prompts. Frozen self-supervised dense features provide visual correspondence rather than object identity: tongue, lip, skin, and boundary patches can all produce high local similarity. We present \method, a retrieval-to-prompt framework that converts a retrieved expert mask into a transferable foreground/background contrast signal for automatic SAM2 point prompting. For each query, frozen DINOv3 descriptors retrieve candidate exemplars from a fixed labeled memory. The corresponding expert mask partitions reference patch features into foreground and background sets, whose dense similarities are transferred to the query as competing correspondence maps. A query-side separability reranker selects the most transferable exemplar, and contrastive foreground-minus-background ranking selects three positive point prompts for SAM2. The method uses a small labeled memory while keeping all model parameters fixed at inference time. We evaluate on HIT-Tongue and SM-Tongue, a 2,155-image smartphone benchmark with expert masks. \method obtains 0.984 mIoU on HIT-Tongue and 0.973 mIoU on SM-Tongue. On the latter, it improves over Tongue-SAM by 3.0 mIoU points and exceeds controlled SAM2 box and center-point prompting. Swapping only the labeled memory also transfers the method across imaging domains, where a U-Net retrained on the full source domain collapses. The results indicate that mask-conditioned retrieval can convert foundation-model features into stable automatic prompts under unconstrained capture conditions.
\end{abstract}

\begin{links}
    \link{Code}{https://github.com/jw-chae/memory-sam}
\end{links}

\section{Introduction}
\label{sec:intro}

Accurate tongue-region segmentation is a prerequisite for quantitative analysis of tongue color, texture, shape, and surface patterns. Errors near the boundary contaminate downstream descriptors by mixing the tongue with lips, skin, teeth, or background, while missing peripheral regions can remove clinically relevant texture and geometry. This requirement is particularly difficult for smartphone images, where illumination, camera distance, viewpoint, facial context, tongue pose, and visible scale vary substantially~\cite{zeng2020bghnet,zhou2019tonguenet}.

Early tongue-segmentation systems relied on deformable contours, shortest paths, and handcrafted gradient priors~\cite{pang2005,sheng2009,shi2013,shi2014,ning2012}. Fully supervised convolutional networks later produced large accuracy gains through learned representations and task-specific losses~\cite{ronneberger2015,long2015,qin2020u2,cai2020,xu2020,badura2021}. These models remain a strong choice when target-domain masks and retraining resources are abundant. Their deployment cost, however, grows whenever acquisition devices, populations, framing protocols, or illumination conditions change, because the segmentation model must be optimized and validated again for the new distribution.

Promptable foundation models offer a different interface. SAM and SAM2 can produce high-quality masks from points, boxes, or masks without training a new decoder for every object category~\cite{kirillov2023,ravi2024}. The interface does not by itself make inference automatic: segmentation quality depends on the source and placement of the prompt. Manual clicks are incompatible with high-throughput screening, and a detector introduces another trained component. For a protruding, non-convex tongue, a rectangular detector box often contains substantial lip, skin, teeth, and facial context. Tongue-SAM~\cite{cao2023tonguesam} demonstrates the practicality of detector-to-box prompting, but also exposes this geometric mismatch. We refer to the resulting requirement for reliable prompts despite a capable mask decoder as the \emph{automation gap}.

This work asks whether a small labeled exemplar memory can close that gap without task-specific parameter optimization. A retrieved exemplar provides more information than a coarse box. Its mask partitions reference features into foreground and background sets, enabling two competing correspondence maps on the query. Foreground similarity suggests where the tongue may be, while background similarity identifies locations that are also compatible with surrounding facial regions. Their difference supplies a direct criterion for choosing discriminative positive prompts.

The idea is simple, but three difficulties make naive nearest-neighbor transfer unreliable. First, the globally nearest image need not provide the best mask-conditioned correspondence because global similarity may be dominated by background appearance. Second, foreground and background similarities can both be high near boundaries and visually ambiguous lip or skin regions. Third, negative points that appear reasonable in feature space can suppress valid tongue regions when passed to SAM2. A usable retrieval-to-prompt system therefore needs to select a transferable exemplar and rank positive points by explicit foreground/background contrast.

\begin{figure*}[t]
    \centering
    \includegraphics[width=0.97\textwidth]{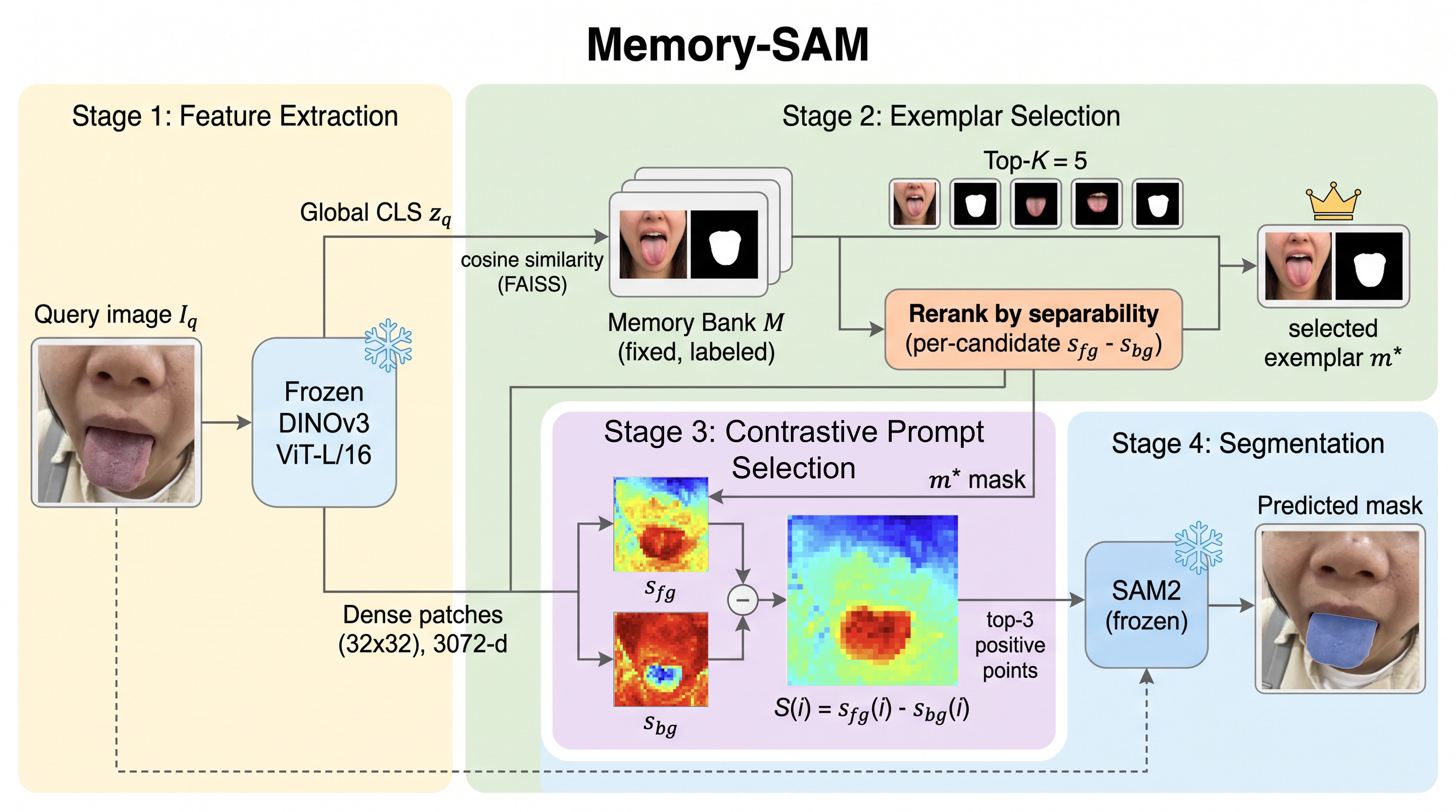}
    \caption{Overview of \method. A query is matched to a fixed labeled exemplar bank using DINOv3 descriptors. The selected exemplar mask partitions reference patches into foreground and background sets. Their transferred similarities are reranked, and the top three foreground-minus-background points prompt SAM2. Test images never enter the memory bank.}
    \label{fig:overview}
\end{figure*}

We propose \method, shown in Fig.~\ref{fig:overview}. A frozen DINOv3 encoder~\cite{simeoni2025dinov3} extracts global and dense descriptors. Global cosine search retrieves five labeled candidates from a fixed training-split memory. Each candidate mask separates its patch descriptors into foreground and background sets; the candidate that induces the clearest query-side separation is selected. Finally, the three locations with the largest foreground-minus-background scores prompt a frozen SAM2 model. The bank never receives test images and is not updated online.

Our contributions are:
\begin{itemize}
\item \textbf{A mask-conditioned contrast mechanism that turns annotation into automatic prompts.} A retrieved expert mask partitions reference features into foreground and background sets, and their query-side difference, $s_{\fg}-s_{\bg}$, selects locations that respond to the target but not to visually similar surroundings. Thus, prompt generation relies on object-specific contrast rather than raw dense similarity.
\item \textbf{A human-prompt-free realization of this mechanism.} Global retrieval, query-side exemplar reranking, and contrastive top-$K$ prompt selection convert a fixed labeled memory into automatic SAM2 point prompts using frozen DINOv3 and SAM2 encoders.
\item \textbf{Evaluation under unconstrained capture.} We benchmark on the public HIT-Tongue dataset and SM-Tongue, a 2,155-image smartphone dataset with expert masks, against supervised, Tongue-SAM, and controlled SAM2 prompt baselines, with runtime, component, memory-size, and cross-domain memory-transfer analyses.
\end{itemize}

\section{Related Work}
\label{sec:related}

\paragraph{Tongue segmentation.}
Classical tongue-segmentation methods separate the tongue from surrounding facial regions using shape priors, active contours, gradient vector flow, and shortest-path formulations~\cite{pang2005,sheng2009,shi2013,shi2014,ning2012}. Deep models later replaced handcrafted boundary cues with learned representations and task-specific losses. Representative systems include two-stage convolutional models, U-Net variants, boundary-aware networks, and joint segmentation-classification objectives~\cite{wang2018,zeng2020bghnet,zhou2019tonguenet,cai2020,xu2020,badura2021}. These methods establish strong supervised baselines, but they optimize a task-specific segmentation network for the target distribution. Our objective is complementary: we keep inference automatic while avoiding target-specific parameter optimization.

\paragraph{Promptable segmentation and medical adaptation.}
SAM introduced a general prompt-conditioned mask decoder, and SAM2 extended this framework to images and videos~\cite{kirillov2023,ravi2024}. Because their outputs depend on prompts, using these models in medical or clinical workflows still requires either human interaction, a prompt generator, or adaptation to the target domain. Medical SAM variants address this issue mainly by modifying or adapting the segmenter, for example through large-scale medical fine-tuning, 2D medical prompt training, lightweight adapters, or SAM2-based medical extensions~\cite{ma2024segment,cheng2023sammed2d,zhu2024medical,chen2024sam2adapter,ma2025medsam2}. These works improve representation or decoder transfer. In contrast, \method keeps both the feature extractor and SAM2 frozen and focuses on the prompt-generation problem: how to obtain reliable prompts automatically from a small labeled memory.

\paragraph{Automatic prompting for SAM.}
A direct way to make SAM automatic is to predict prompts with an auxiliary model. Detector-to-SAM pipelines generate bounding boxes and then rely on the promptable decoder to produce the final mask. Tongue-SAM follows this strategy for tongue images by integrating an object-detection prompt generator with SAM~\cite{cao2023tonguesam}. Such systems are practical, but a rectangular box is a coarse geometric prompt. For a protruding and non-convex tongue, the box can include lips, teeth, skin, and other facial context, making the prompt ambiguous even when the mask decoder is strong. Other self-prompting methods generate boxes, masks, or points through learned auxiliary prompt generators~\cite{xie2025selfpromptsam}. Our method instead derives sparse positive prompts from mask-conditioned foreground/background evidence, without training a detector or prompt generator.

\paragraph{Reference-based and retrieval-augmented segmentation.}
Several works use a reference image or a small support set to guide foundation-model segmentation. PerSAM personalizes SAM from a single reference image and mask, while Matcher uses all-purpose feature matching to perform one-shot segmentation with off-the-shelf foundation models~\cite{zhang2024persam,liu2024matcher}. Medical variants such as P$^2$SAM and Med-PerSAM further explore patient-specific or one-shot prompting, often selecting multiple points or using warping-based prompt refinement~\cite{zhao2024p2sam,yoon2024medpersam}. Retrieval-augmented few-shot medical segmentation has also been studied by retrieving similar annotated samples with DINOv2 and using them as conditions for SAM2 memory attention~\cite{zhao2024retrieval}. \method differs in how the retrieved annotation is used. We do not warp the support mask to the query, personalize SAM, fine-tune prompt weights, or encode retrieved samples as SAM2 memories. The retrieved expert mask is used only to define foreground and background feature sets, whose query-side contrast determines a small set of SAM2 point prompts.

\paragraph{Self-supervised dense correspondence and contrastive prompting.}
Self-supervised ViTs such as DINOv2 and DINOv3 provide transferable global and dense patch descriptors, and efficient nearest-neighbor search makes fixed-bank retrieval practical at inference time~\cite{oquab2024,simeoni2025dinov3,johnson2017,douze2024}. However, dense feature similarity provides visual correspondence rather than object identity. In tongue images, patches from the tongue boundary, lips, teeth, and illuminated skin can be locally similar, so raw foreground matching alone can produce diffuse or unstable prompt candidates. \method addresses this limitation by converting a retrieved expert annotation into a foreground/background contrastive signal: locations are selected when $s_{\fg}$ is high and competing $s_{\bg}$ is low. Thus, our contribution is not retrieval itself, but the conversion of retrieved annotation into object-specific prompt evidence for SAM2.

\section{Method}
\label{sec:method}

\subsection{Problem Formulation and Fixed Memory}

Let $I_q$ be a query image and $Y_q$ its unknown binary tongue mask. The objective is to predict $\hat Y_q$ without any prompt supplied by a user. The fixed memory bank is
\begin{equation}
    \mathcal{M}=\{(I_m,Y_m)\}_{m=1}^{M},
\end{equation}
where each $I_m$ is a training-split image and $Y_m$ is its expert mask. The bank is created once, contains no test image, and remains unchanged during evaluation.

All images are resized to $512\times512$ by direct square resizing. We use a frozen DINOv3 ViT-L/16 with the LVD-1689M pretrained checkpoint (\texttt{dinov3\_vitl16\_pretrain\_lvd1689m}). The patch/16 tokenizer yields a $32\times32$ dense grid. Dense patch descriptors are obtained by concatenating the token features of layers 11, 17, and 23 (the last), after removing the CLS and register tokens, giving $3072$-dimensional descriptors $Q=\{q_i\}_{i=1}^{1024}$. The global retrieval descriptor $z_q$ is the last-layer CLS token. For every memory image, we precompute the global descriptor $z_m$, dense grid $R_m=\{r_j^{(m)}\}$, and its mask on the feature grid: the exemplar mask is resized to $32\times32$ by nearest-neighbor interpolation and binarized at $127$, so patches with positive mask value define the foreground set and the rest define the background set. Global and patch descriptors are $\ell_2$-normalized, making cosine similarity equivalent to an inner product.

This design separates representation reuse from task adaptation. The foundation encoders are fixed, while the small memory supplies task identity and spatial supervision. Adding or replacing a memory exemplar changes the available evidence but does not change model parameters.

\subsection{Global Retrieval and Query-Side Reranking}

The global descriptor retrieves a shortlist rather than committing to a single neighbor. Using exact inner-product search with FAISS~\cite{johnson2017,douze2024}, we obtain
\begin{equation}
    \mathcal{N}_q=\TopK_{m\in\mathcal{M}}\langle z_q,z_m\rangle,
    \qquad |\mathcal{N}_q|=K,
\end{equation}
with $K=5$ by default. Global retrieval captures broad pose, framing, and appearance similarity, but it does not know whether the stored mask will induce useful local correspondences. We therefore score every shortlisted candidate on the query.

For candidate $m$, the downsampled mask partitions its normalized descriptors into
\begin{align}
R_{\fg}^{(m)} &= \{r_j^{(m)}:Y_m(j)=1\}, \\
R_{\bg}^{(m)} &= \{r_j^{(m)}:Y_m(j)=0\}.
\end{align}
For every query patch $q_i$, we compute its best match to each partition:
\begin{align}
 s_{\fg}^{(m)}(i) &= \max_{r\in R_{\fg}^{(m)}}\langle q_i,r\rangle, \\
 s_{\bg}^{(m)}(i) &= \max_{r\in R_{\bg}^{(m)}}\langle q_i,r\rangle.
\label{eq:similarity}
\end{align}
The two maps answer different questions: whether a query location resembles any annotated tongue patch and whether it also resembles the exemplar's surrounding context. A candidate is useful when it creates a strong foreground response that is not matched by an equally strong background response. We select
\begin{equation}
 m^*=\arg\max_{m\in\mathcal{N}_q}
 \left[\max_i s_{\fg}^{(m)}(i)-\max_i s_{\bg}^{(m)}(i)\right].
\label{eq:rerank}
\end{equation}
Unlike global top-1 selection, this reranker explicitly evaluates transferability on the current query. It is parameter-free and requires only the dense similarities already needed for prompt construction.

\subsection{Contrastive Prompt Selection}

For the selected exemplar, every query patch is scored by the contrastive difference between its foreground and background responses:
\begin{equation}
    \mathcal{S}(i)=s_{\fg}(i)-s_{\bg}(i).
\label{eq:contrast}
\end{equation}
This score is high only where a location resembles annotated tongue and, at the same time, is unlike the exemplar's surrounding context. We rank all $32\times32$ query patches by $\mathcal{S}(i)$ and take the top $K_{\fg}=3$. A selected patch at grid row $r$ and column $c$ maps to image coordinate $\bigl((c+0.5)\cdot\tfrac{W}{32},\,(r+0.5)\cdot\tfrac{H}{32}\bigr)$, i.e.\ the center of its $16\times16$ patch at $512\times512$ input, and is passed to SAM2 as a positive point prompt. Ranking over the full grid means the prompt set is never empty and requires no threshold. Although $s_{\bg}$ is essential to the ranking, background points are not passed to SAM2 as explicit negatives; our ablation shows that retaining them causes severe under-segmentation.

Algorithm~\ref{alg:memorysam} summarizes the complete inference procedure. Its only dataset-specific state is the labeled memory; no gradient, optimizer, or pseudo-label update is used at test time.

\begin{algorithm}[t]
\caption{Memory-SAM Inference}
\label{alg:memorysam}
\textbf{Input}: query $I_q$, fixed bank $\mathcal{M}$, frozen DINOv3 and SAM2\\
\textbf{Output}: predicted mask $\hat Y_q$
\begin{algorithmic}[1]
\STATE Extract normalized global and patch descriptors $(z_q,Q)$.
\STATE Retrieve $\mathcal{N}_q$, the top-$K$ memory images by $\langle z_q,z_m\rangle$.
\FOR{each $m\in\mathcal{N}_q$}
\STATE Partition $R_m$ into $R_{\fg}^{(m)}$ and $R_{\bg}^{(m)}$ with $Y_m$.
\STATE Compute $s_{\fg}^{(m)}$ and $s_{\bg}^{(m)}$ using Eq.~\ref{eq:similarity}.
\ENDFOR
\STATE Select $m^*$ using query-side separability in Eq.~\ref{eq:rerank}.
\STATE Rank all query patches by $\mathcal{S}=s_{\fg}-s_{\bg}$ (Eq.~\ref{eq:contrast}) and take the top three.
\STATE Map the top three locations to positive image-space points.
\STATE \textbf{return} SAM2$(I_q,\text{points})$.
\end{algorithmic}
\end{algorithm}

\begin{figure*}[t]
    \centering
    \includegraphics[width=0.98\textwidth]{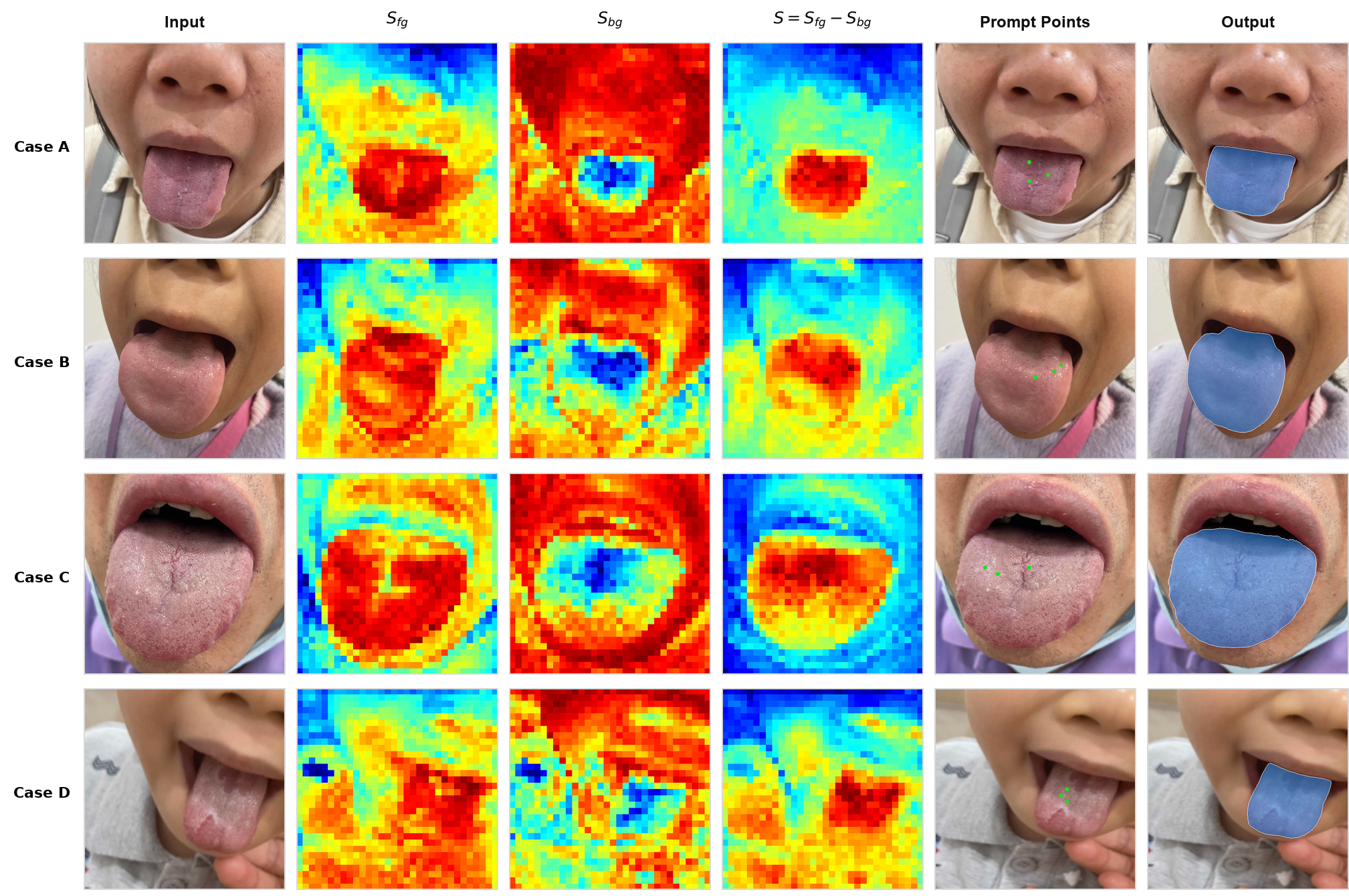}
    \caption{Prompt formation on representative smartphone images. Foreground similarity alone can remain diffuse, while background similarity identifies facial context. Their difference concentrates evidence on the tongue; the selected positive points then produce the final SAM2 mask.}
    \label{fig:similarity}
\end{figure*}

\subsection{Design Rationale}

The central operation of \method is the conversion of a retrieved mask annotation into a contrastive prompt signal. A frozen self-supervised encoder yields descriptors that capture visual correspondence, but visual similarity is not object identity: tongue boundaries, lips, teeth, and adjacent skin can match annotated tongue patches as strongly as the tongue interior does. Consequently, the foreground similarity map $s_{\fg}$ alone can remain diffuse, and ranking by it alone may produce unstable prompts. The retrieved expert mask resolves this ambiguity by also defining a background set, giving a second map $s_{\bg}$ that measures resemblance to surrounding context. Their difference, $s_{\fg}-s_{\bg}$, is high only where a query location looks like the target and unlike the surroundings, which is the desired condition for a discriminative positive prompt.

The remaining stages are designed to preserve this contrast under retrieval errors, boundary ambiguity, and prompt sensitivity. Global retrieval first narrows the search to exemplars with compatible pose, framing, and appearance. Query-side reranking then prevents a globally similar but spatially unhelpful reference from dominating the result by selecting the exemplar that induces the clearest foreground/background separation on the query. Finally, contrastive ranking selects only the most reliable high-contrast locations for SAM2.

The method deliberately uses sparse positive prompts rather than transferring the exemplar mask directly. Direct mask warping would require geometric alignment and could copy reference-shape errors to the query. Sparse points instead communicate high-confidence object identity while allowing SAM2 to infer the query-specific boundary from its own image representation. Using multiple points rather than a single positive point is also important for protruding and non-convex tongue regions, where small changes in point location may cause SAM2 to favor only a partial tongue region or nearby facial context. Multiple high-contrast points reduce this prompt ambiguity while still leaving boundary inference to SAM2.

\section{Experiments}
\label{sec:experiments}

\subsection{Experimental Setup}

\paragraph{Datasets and protocol.}
HIT-Tongue contains 300 tongue images\cite{cao2023tonguesam}. SM-Tongue contains 2,155 smartphone-captured images with variation in illumination, viewpoint, distance, facial expression, pose, and background. Three certified practitioners jointly reviewed the images and expert masks. Each dataset is independently split 70/30 using seed 42, yielding 210/90 HIT-Tongue and 1,508/647 SM-Tongue train/test images. Supervised models are fitted on the corresponding training split. \method constructs its candidate memory exclusively from that split, and no test image is used for bank construction, retrieval targets, or online updating.

The human-subject collection was approved by the responsible institutional review board, all participants provided written informed consent, and the study data were anonymized. Paper figures are cropped to the task-relevant lower-face region. We report mean intersection over union (mIoU), mean pixel accuracy (mPA), overall pixel accuracy, precision, recall, and Dice. Here mIoU and mPA are averaged over the foreground and background classes, while precision, recall, and Dice are computed on the foreground tongue class. All metrics are computed at $512\times512$ resolution, with masks resized by nearest-neighbor interpolation.

\paragraph{Baselines.}
We compare with U-Net~\cite{ronneberger2015}, FCN~\cite{long2015}, and U$^2$-Net~\cite{qin2020u2}, which represent fully supervised segmentation. These supervised baselines use $512\times512$ inputs, standard augmentation, Adam with learning rate $10^{-4}$, at most 100 epochs, and early stopping. Tongue-SAM~\cite{cao2023tonguesam} supplies detector-generated boxes; to isolate prompt generation rather than mask-decoder capacity, its original SAM module is replaced with the same SAM2.1 Hiera-L checkpoint used by \method. We additionally report SAM2 with one ground-truth-derived box or one ground-truth-derived center point per image. These rows serve as controlled prompt-type references rather than human-prompt-free baselines. The box is defined by the mask bounding rectangle with a 20-pixel margin from the mask extremities, and the point prompt is placed at the mask centroid.

\paragraph{Implementation.}
Unless otherwise stated, the fixed bank contains $M=20$ exemplars randomly selected from the training split using seed 0. The global search returns $K=5$ candidates, and the top three contrastive patches are retained as positive prompts. Masks are produced by SAM2.1 Hiera-L with multimask output enabled, keeping the candidate with the highest predicted IoU. All experiments use a single NVIDIA RTX 3090 GPU. End-to-end \method inference includes DINOv3 feature extraction, retrieval, reranking, prompt construction, and SAM2 decoding.

\subsection{Main Results}

Tables~\ref{tab:hit} and~\ref{tab:sm} present the complete metrics. On HIT-Tongue, where supervised baselines are near ceiling, \method reaches 0.984 mIoU and 0.987 Dice, on par with FCN and ahead of Tongue-SAM and center-point SAM2. Its mIoU is also 0.2 points above controlled box prompting. The result shows that a small fixed memory can produce automatic prompts comparable to strong controlled SAM2 prompts in a controlled image setting.

On SM-Tongue, the distinction is clearer. \method obtains 0.973 mIoU and 0.977 Dice, improving over Tongue-SAM by 3.0 and 3.2 points, respectively. It exceeds SAM2 with a box by 2.8 mIoU points and SAM2 with a center point by 0.6 points. The cross-dataset change also highlights prompt robustness: from HIT-Tongue to SM-Tongue, \method drops by 1.1 mIoU points, whereas box-prompt SAM2 and Tongue-SAM each drop by 3.7 points.

\begin{table}[t]
\centering
\small
\setlength{\tabcolsep}{1.05mm}
\begin{tabular}{@{}lcccccc@{}}
\toprule
Method & mIoU & mPA & Acc. & Prec. & Rec. & Dice \\
\midrule
U-Net & \textbf{.990} & \textbf{.994} & \textbf{.996} & \textbf{.992} & \textbf{.992} & \textbf{.992} \\
U$^2$-Net & .988 & \textbf{.994} & \textbf{.996} & \textbf{.992} & .988 & .990 \\
FCN & .984 & .991 & .994 & .990 & .985 & .987 \\
\midrule
SAM2 + box & .982 & .986 & .988 & .987 & .982 & .984 \\
SAM2 + center & .978 & .981 & .984 & .983 & .977 & .980 \\
Tongue-SAM & .980 & .983 & .986 & .985 & .979 & .982 \\
\method(ours) & \underline{.984} & \underline{.990} & \underline{.991} & \underline{.989} & \underline{.985} & \underline{.987} \\
\bottomrule
\end{tabular}
\caption{Results on HIT-Tongue ($n=90$). Bold denotes the best overall result; underline denotes the best SAM-based result. Box and center prompts are ground-truth-derived controlled prompts.}
\label{tab:hit}
\end{table}

\begin{table}[t]
\centering
\small
\setlength{\tabcolsep}{1.05mm}
\begin{tabular}{@{}lcccccc@{}}
\toprule
Method & mIoU & mPA & Acc. & Prec. & Rec. & Dice \\
\midrule
U-Net & \textbf{.983} & \textbf{.991} & \textbf{.995} & \textbf{.989} & \textbf{.985} & \textbf{.987} \\
FCN & .981 & .990 & .994 & .987 & .983 & .985 \\
U$^2$-Net & .978 & .989 & .993 & .983 & .982 & .982 \\
\midrule
SAM2 + box & .945 & .963 & .978 & .959 & .936 & .947 \\
SAM2 + center & .967 & .982 & .987 & .975 & .967 & .971 \\
Tongue-SAM & .943 & .961 & .977 & .958 & .933 & .945 \\
\method(ours) & \underline{.973} & \underline{.987} & \underline{.990} & \underline{.979} & \underline{.976} & \underline{.977} \\
\bottomrule
\end{tabular}
\caption{Results on SM-Tongue ($n=647$). Formatting follows Table~\ref{tab:hit}.}
\label{tab:sm}
\end{table}

\begin{figure*}[t]
    \centering
    \includegraphics[width=0.50\textwidth]{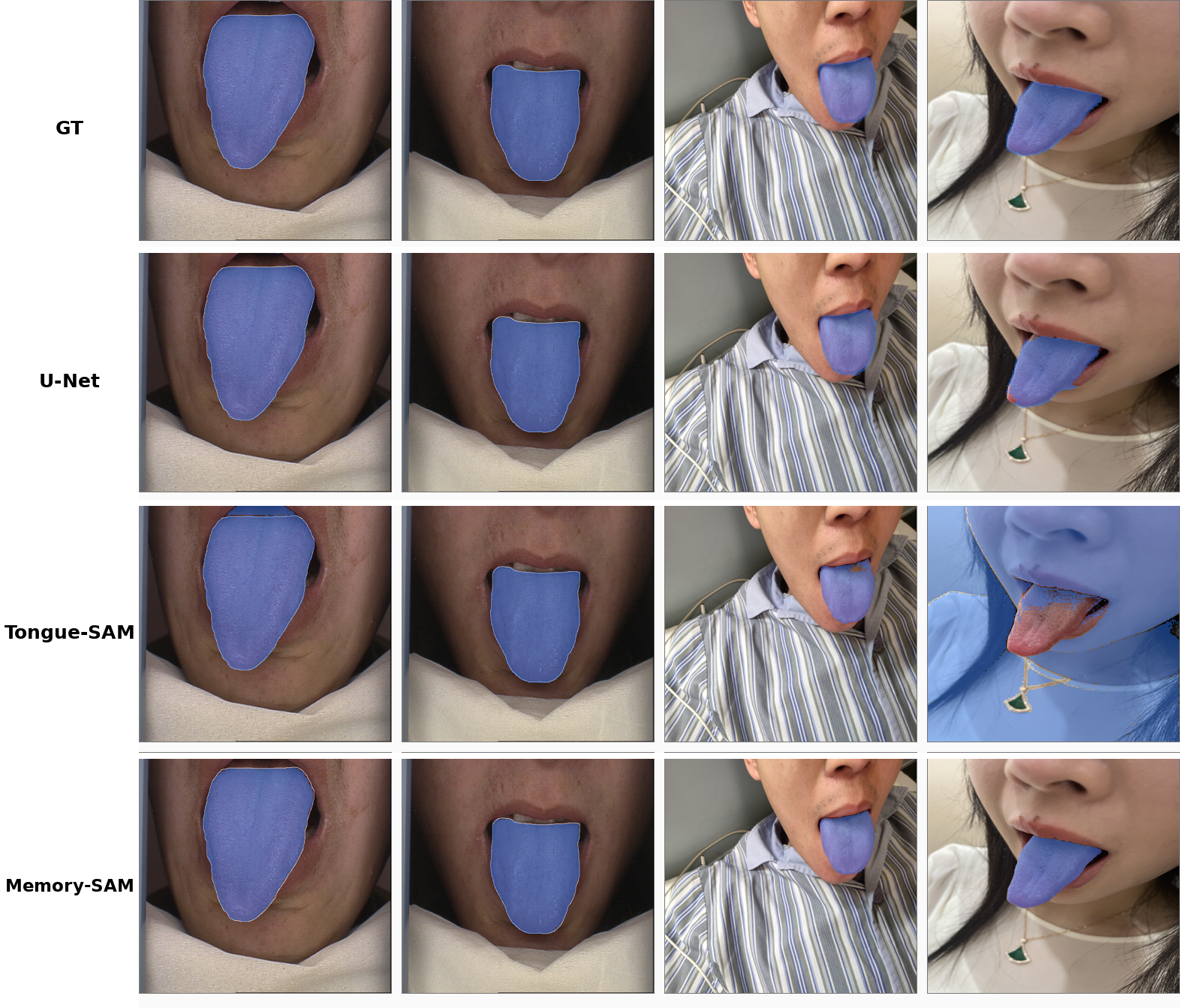}
    \caption{Qualitative comparison. The first two columns show controlled HIT-Tongue examples and the last two show unconstrained SM-Tongue examples. Blue overlays denote predicted or reference masks. Memory-guided points reduce the facial spillover observed with detector-to-box prompting.}
    \label{fig:qualitative}
\end{figure*}

\paragraph{Qualitative behavior.}
Figure~\ref{fig:qualitative} contrasts controlled and unconstrained examples. U-Net remains accurate after target-specific training. Tongue-SAM can also succeed when its box tightly surrounds the object, but a loose box may cause the mask to spill into adjacent facial regions. \method instead places points where retrieved foreground evidence dominates background evidence, producing tighter masks without a detector or manual prompt. The examples also clarify that the memory does not transfer a fixed tongue shape: SAM2 adapts the final boundary to each query after receiving only sparse positive points.

\paragraph{Efficiency and memory behavior.}
End-to-end inference takes 626 ms/image on an RTX 3090 at $512\times512$, including feature extraction, top-5 retrieval and reranking, prompt construction, and SAM2 inference. The bank is fixed and small: only global descriptors participate in the first search, and dense correspondence is evaluated for the shortlisted candidates. One selected exemplar ultimately determines the prompt set. This organization avoids running SAM2 once per candidate and keeps retrieval overhead bounded by $K$ rather than the full bank size.

\subsection{Ablation Studies}
\label{sec:ablation}

Table~\ref{tab:ablation} reports ablations on SM-Tongue, and together they confirm that foreground/background contrast, not dense similarity alone, is what makes the prompts reliable. Ranking query patches by raw foreground similarity instead of the contrastive score $\mathcal{S}=s_{\fg}-s_{\bg}$ loses 1.0 mIoU point, so a location's resemblance to a tongue patch is not sufficient unless it is also unlike the exemplar's surroundings. The largest degradation occurs when background points are added to the SAM2 prompt as explicit negatives (the three most background-like patches, ranked by $s_{\bg}-s_{\fg}$): mIoU falls from 0.973 to 0.923. Background evidence thus contributes most as a \emph{selection penalty} rather than as an explicit negative instruction to the mask decoder.

The selection stages protect this contrast. Replacing the full procedure with global top-1 retrieval loses 1.1 mIoU points, showing that the globally closest image is not always the most transferable. Reducing the reranking pool from five candidates to three loses 0.5 points.

Memory diversity also matters. A single exemplar reaches 0.893 mIoU, while 5, 10, and 20 exemplars reach 0.950, 0.954, and 0.973. The bank handles query variation through exemplar diversity rather than parameter updates.

\begin{table}[t]
\centering
\small
\setlength{\tabcolsep}{1.5mm}
\begin{tabular}{@{}lrrrr@{}}
\toprule
Setting & mIoU & $\Delta$ & Dice & $\Delta$ \\
\midrule
\textbf{Full} & \textbf{.973} & .000 & \textbf{.977} & .000 \\
Top-1 global only & .962 & -.011 & .969 & -.008 \\
Top-3 retrieved exemplars & .968 & -.005 & .973 & -.004 \\
Rank by $s_{\fg}$ only & .963 & -.010 & .966 & -.011 \\
Keep background prompts & .923 & -.050 & .948 & -.029 \\
\midrule
$M=1$ & .893 & -- & .904 & -- \\
$M=2$ & .915 & -- & .927 & -- \\
$M=5$ & .950 & -- & .957 & -- \\
$M=10$ & .954 & -- & .960 & -- \\
$M=20$ & \textbf{.973} & -- & \textbf{.977} & -- \\
\bottomrule
\end{tabular}
\caption{Ablations on SM-Tongue ($n=647$). The upper block changes one component of the default configuration; the lower block varies fixed-bank size.}
\label{tab:ablation}
\end{table}

\subsection{Cross-Domain Memory Transfer}
\label{sec:cross}

Because \method never updates model parameters, adapting to a new imaging domain requires only swapping the labeled memory. We test this directly by evaluating each test set with a memory built from the \emph{other} dataset (same $M=20$ exemplars, seed 0, identical inference), and we contrast it with a supervised U-Net that is retrained on the source domain. To separate the effect of the paradigm from the effect of data volume, U-Net is trained both on the full source split and on a matched 20-image subset. Table~\ref{tab:cross} reports all transfers, grouped by target domain, with the in-domain \method rows from Tables~\ref{tab:hit} and~\ref{tab:sm} as references.

The paradigm gap is large and one-sided. Trained on the full HIT split, U-Net still collapses to 0.585 mIoU on smartphone queries, whereas \method retains 0.922 mIoU using only a 20-image HIT memory and no retraining. The reverse direction is milder because the smartphone set is larger and more varied, but the ordering is the same: full-train U-Net reaches 0.910 mIoU on HIT while \method reaches 0.934. Matching the data budget makes the contrast sharper still, as 20-image U-Net drops to 0.756 and 0.493, respectively. The transfer profile of \method is also benign: with a clinical memory on smartphone queries, recall stays at 0.991 while precision falls to 0.881, so the contrastive points still land on the tongue and only the boundary spills slightly into facial context. End-to-end latency is unchanged at 626~ms/image, since only the memory contents differ. A retrained segmenter cannot transfer this way: it optimizes to the source distribution, whereas \method reuses frozen encoders and swaps a small labeled memory.

\begin{table}[t]
\centering
\small
\setlength{\tabcolsep}{1.4mm}
\begin{tabular}{@{}llcc@{}}
\toprule
Method & Source (train / memory) & mIoU & Dice \\
\midrule
\multicolumn{4}{@{}l}{\emph{Target: HIT-test ($n=90$)}}\\
\method & HIT-20 memory (in-domain)$^\dagger$ & .984 & .987 \\
\method & SM-20 memory (no retrain) & \textbf{.934} & \textbf{.943} \\
U-Net & SM-full train (1{,}508) & .910 & .925 \\
U-Net & SM-20 train & .756 & .773 \\
\midrule
\multicolumn{4}{@{}l}{\emph{Target: SM-test ($n=647$)}}\\
\method & SM-20 memory (in-domain)$^\dagger$ & .973 & .977 \\
\method & HIT-20 memory (no retrain) & \textbf{.922} & \textbf{.933} \\
U-Net & HIT-full train (210) & .585 & .491 \\
U-Net & HIT-20 train & .493 & .285 \\
\bottomrule
\end{tabular}
\caption{Cross-domain transfer (source $\neq$ target). \method swaps a 20-image memory with no parameter update; U-Net is retrained on the source domain at full and matched-20 budgets. $^\dagger$In-domain \method references from Tables~\ref{tab:hit} and~\ref{tab:sm}.}
\label{tab:cross}
\end{table}

\section{Conclusion}
\label{sec:conclusion}

We introduced \method, a retrieval-to-prompt framework that generates automatic SAM2 point prompts from a small labeled exemplar memory. The key idea is to use a retrieved mask not as a shape template, but as a source of foreground/background evidence for selecting reliable query-side prompts. On HIT-Tongue and SM-Tongue, this design improves robustness over detector-to-box prompting without task-specific parameter optimization. It also transfers across imaging domains by swapping the labeled memory alone: a supervised U-Net retrained on the full source domain collapses on the target, while \method holds its accuracy with a 20-image memory and no retraining. The results suggest that labeled-memory prompting can be a practical bridge between fully supervised segmentation and interactive foundation-model segmentation. Although the contrast mechanism is not tied to tongue anatomy, this study validates it only for tongue segmentation; extension to structures with different shape, texture, and boundary statistics remains future work. Future work will also emphasize memory construction, multi-site and multi-device validation, and efficient deployment.
\bibliography{main}

\end{document}